\documentclass[runningheads]{llncs}
\usepackage[T1]{fontenc}
\usepackage{graphicx}
\usepackage{multirow}
\usepackage{booktabs}
\usepackage{amssymb}
\usepackage[colorlinks=true,citecolor=blue,urlcolor=blue]{hyperref}
\usepackage[misc]{ifsym} 
\usepackage{subfigure}

\begin{document}
	\title{FedIIC: Towards Robust Federated Learning for Class-Imbalanced Medical Image Classification}
	\titlerunning{Federated Learning for Class-Imbalanced Medical Image Classification}
	\author{
		Nannan Wu\inst{1} %index{Wu, Nannan}
		\and
		Li Yu\inst{1} %index{Yu, Li}
		\and
		Xin Yang\inst{1} %index{Yang, Xin}
		\and 
		Kwang-Ting Cheng\inst{2} %index{Cheng, Kwang-Ting}
		\and 
		Zengqiang Yan\textsuperscript{1(\Letter)} %index{Yan, Zengqiang}
	}
	\authorrunning{N. Wu et al.}
	\institute{School of Electronic Information and Communications, Huazhong University of Science and Technology \\
		\email{\{wnn2000,hustlyu,xinyang2014,z\_yan\}@hust.edu.cn} 
		\and School of Engineering, Hong Kong University of Science and Technology 
		\email{timcheng@ust.hk}
	}
	\maketitle              % typeset the header of the contribution
	\begin{abstract}
		Federated learning (FL), training deep models from decentralized data without privacy leakage, has shown great potential in medical image computing recently. However, considering the ubiquitous class imbalance in medical data, FL can exhibit performance degradation, especially for minority classes (\textit{e.g.} rare diseases). Existing methods towards this problem mainly focus on training a  balanced classifier to eliminate class prior bias among classes, but neglect to explore better representation to facilitate classification performance. In this paper, we present a privacy-preserving FL method named FedIIC to combat class imbalance from two perspectives: feature learning and classifier learning. In feature learning, two levels of contrastive learning are designed to extract better class-specific features with imbalanced data in FL. In classifier learning, per-class margins are dynamically set according to real-time difficulty and class priors, which helps the model learn classes equally. Experimental results on publicly-available datasets demonstrate the superior performance of FedIIC in dealing with both real-world and simulated multi-source medical imaging data under class imbalance. Code is available at \url{https://github.com/wnn2000/FedIIC}.
		\keywords{Federated learning \and Class imbalance \and Contrastive learning \and Classification.}
	\end{abstract}
	%x
	%
	%
	\section{Introduction}
	
	Federated learning (FL), allowing decentralized data sources to train a unified deep learning model collaboratively without data sharing, has drawn great attention in medical imaging due to its privacy-preserving properties \cite{li2021fedbn,yan2020variation,liu2021feddg,jiang2022harmofl}. Existing studies of FL mainly focus on data heterogeneity across clients \cite{li2020federated,li2021model,mu2023fedproc}, while ignoring the widely-existed class imbalance problem in medical scenarios. In clinical practice, the number of samples for different diseases may vary greatly due to varying incidence rates in the population. When conducting FL on cooperative medical institutions with global class-imbalanced data, the global model may suffer from significant performance degradation, which typically manifests as the recognition accuracy of minority classes (\textit{e.g.} rare diseases) being lower than that of majority classes (\textit{e.g.} common diseases) \cite{shang2022federated}. Deploying such a biased global/federated model is fatal, especially for misdiagnosing a rare disease \cite{ju2022flexible,yang2022proco}. Therefore, addressing class imbalance in federated learning is of great value.
	
	Several FL frameworks have been proposed to tackle imbalanced data \cite{yang2021federated,duan2020self}. Following re-weighting \cite{cui2019class}, Wang \textit{et al.} \cite{wang2021addressing} presented a weighted form of cross entropy loss named ratio loss depending on a balanced auxiliary dataset for the server to calculate weights. Sarkar \textit{et al.} \cite{sarkar2020fed} introduced focal loss \cite{lin2017focal} to up-weight hard samples. CLIMB \cite{shen2022agnostic} assigned larger weights to clients more likely to own minority classes via a meta-algorithm. Inspired by decoupling \cite{kang2020decoupling}, CReFF \cite{shang2022federated} retrained a new classifier with balanced synthetic features in the server. All these methods aim to balance classes from the classifier perspective without exploring better representations with class-imbalanced data for performance improvement.
	
	In this paper, we formulate the effect of class imbalance in FL into the attribute bias and the class bias \cite{tang2022invariant}. The attribute bias means minority classes have more imbalanced background attributes in their class-specific attributes compared to majority classes, making them less distinguishable. The class bias represents the difference in prior probabilities across classes, resulting in biased predictions toward majority classes. To handle the two biases, we present a new class-balancing FL method named \textbf{FedIIC} from two perspectives: feature learning and classifier learning. The key idea of FedIIC is to alleviate the two biases through the calibration of the feature extractor and the classifier. Specifically, two-level supervised contrastive learning \cite{khosla2020supervised}, \textit{i.e.} intra- and inter-client contrastive learning, is built to calibrate the feature extractor for better feature learning. For classifier learning, difficulty-aware logit adjustment is adopted to calibrate the classifier dynamically for better decision boundaries. Extensive comparison experiments on both real-world and simulated multi-source data validate FedIIC's effectiveness.

	The main contributions are summarized as follows. (1) A new viewpoint of realistic medical FL scenarios where global training data is class-imbalanced. (2) A novel privacy-preserving framework FedIIC for balanced federated learning. (3) Superior performance in dealing with class imbalance under both real-world and simulated multi-source decentralized settings.
	
	\section{Methodology}
	
	\subsection{Preliminaries and Overview}
	
	Considering a typical FL scenario for multi-class image classification with $K$ participants, each participant is assumed to own a private dataset $D_k = \{(x_i, y_i)\}_{i=1}^{N_k}$, $k \in [K]$, where $N_k$ is the data amount of $D_k$, and denote each image-label pair as $(x_i \in \mathcal{X} \subseteq \mathbb{R}^d, y_i \in  \mathcal{Y}  = [L])$. The goal of FL is to train a global model $f(g(\cdot))$ with the union of all cooperative data sources $D:=\cup_{k \in [K]} D_k$ without privacy leakage, where $f(\cdot)$ and $g(\cdot)$ represent the linear classifier and the feature extractor respectively. Note that $D$ is set as class-imbalanced in this paper.
	
	\begin{figure}[!t]
		\centering
		\includegraphics[width=1\textwidth]{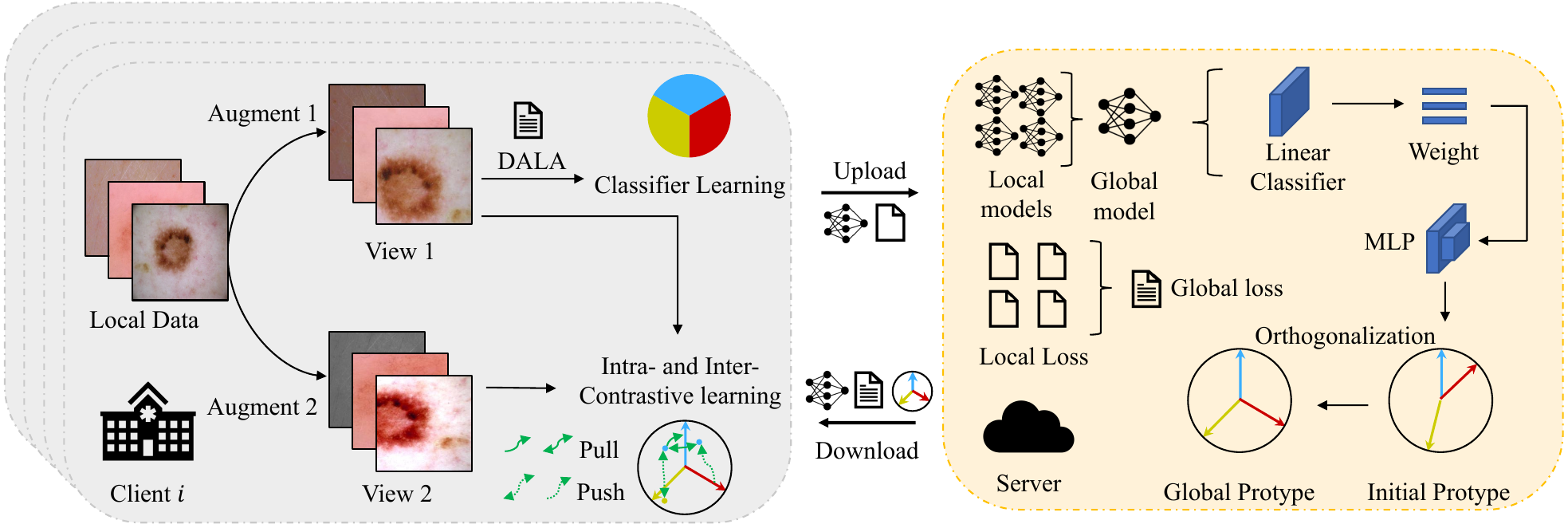}
		\caption{Overview of the proposed FedIIC.}
		\label{framework}
	\end{figure}
	Assuming each image has two kinds of latent attributes, \textit{i.e.} $\mathcal{Z}_c$ and $\mathcal{Z}_a$, representing the class-specific attributes (determining the category of the image, \textit{e.g}. texture, color, \textit{etc.}) and the variant background attributes (\textit{e.g}. brightness, contrast, \textit{etc.}) respectively \cite{tang2022invariant}, based on the Bayes theorem, the posterior probability of classification can be formulated as
	\begin{equation} \label{bayes}
	P(y \mid x) = P(y \mid \mathcal{Z}_c,\mathcal{Z}_a) = \frac{p(\mathcal{Z}_c \mid y)}{p(\mathcal{Z}_c)} \cdot \frac{p(\mathcal{Z}_a \mid y,\mathcal{Z}_c)}{p(\mathcal{Z}_a \mid \mathcal{Z}_c)} \cdot p(y),
	\end{equation}
	where the last two items represent the attribute bias and the class bias respectively, which widely exist in class-imbalanced data and affect the posterior probability. For robust FL with class-imbalanced data, the key idea is to alleviate the two biases simultaneously, instead of focusing on the latter as \cite{shang2022federated}. Hence, we propose FedIIC to address class imbalance from the two perspectives as illustrated in Fig. \ref{framework}. Details are presented in the following.
	
	\subsection{Intra-Client Contrastive Learning}
	
	Limited local data affects data diversity (\textit{i.e.}, limited $(\mathcal{Z}_c, \mathcal{Z}_a)$ combinations), especially for minority classes, making $\mathcal{Z}_c$ less distinguishable. To emphasize more on the learning of $\mathcal{Z}_c$, supervised contrastive learning (SCL), proven to be effective for representation learning \cite{marrakchi2021fighting,kang2021exploring,li2022targeted,zhu2022balanced}, is introduced in local training. The basic loss function of SCL can be formulated as
	\begin{equation} \label{SCL}
	\mathcal{L}_{SCL}=\sum_{i \in I} \frac{-1}{\left| P(i)\right|} \sum_{j \in P(i)} \log \frac{\exp (z_i \cdot z_j / \tau)}{\sum_{a \in A(i)}\exp (z_i \cdot z_a / \tau)},
	\end{equation}
	where $I$ denotes the index set of the multi-view batch generated by different augmentations (\textit{e.g.} the two views in Fig. \ref{framework} ), $\left| \cdot \right|$ measures the number of elements in a set, $A(i) = I \backslash \{i\}, P(i) = \{s \in A(i) \vert y_s=y_i \}$, $\tau$ represents the temperature, and $z$ denotes the $l_2$-normalized embedding of a sample $x$. Note that in this paper, we use a 2-layer MLP $h(\cdot)$ to obtain $z$ before it is normalized as \cite{chen2020simple}, \textit{i.e.} $z=\frac{h(g(x))}{	\left\| h(g(x)) \right\|_2}$. In the multi-view batch, $\mathcal{L}_{SCL}$ keeps the embeddings of the same class closer while pushing the embeddings of different classes further away, which helps the model learn better $\mathcal{Z}_c$ of each class due to richer $\mathcal{Z}_a$.
	However, SCL can not perfectly address class imbalance as the majority classes would benefit more from Eq. \ref{SCL} following traditional training losses (\textit{e.g.} the cross entropy loss). To overcome this problem, we propose to employ a dynamic temperature $\tau^{\prime} := P\tau = (p^i p^j)^t\tau$ in Eq. \ref{SCL} inspired by \cite{zhu2022balanced,kang2021exploring}, where $p^i$ is the prior probability of class $i$ in the local dataset and $t$ is a parameter set as 0.5 by default. Hence, the loss function is rewritten as 
	\begin{equation} \label{intra}
	\mathcal{L}_{Intra}=\sum_{i \in I} \frac{-1}{\left| P(i)\right|} \sum_{j \in P(i)} \log \frac{\exp (z_i \cdot z_j / \tau^{\prime})}{\sum_{a \in A(i)}\exp (z_i \cdot z_a / \tau^{\prime})},
	\end{equation}
	named intra-client contrastive learning. Through $P$, sample pairs of the minority classes are up-weighted compared to those of the majority classes, leading to better balance. 
	
	\subsection{Inter-Client Contrastive Learning}
	Given limited local data under FL, the effectiveness of intra-client contrastive learning may be bounded. How to better utilize cross-client data from the global perspective is crucial for further performance improvement. Inspired by learning from prototypes \cite{mu2023fedproc,chen2022personalized,guo2022dual}, we propose inter-client contrastive learning. Assuming a set of shared class-wise prototypes $V=\{v^1, v^2, ..., v^L\}$ across clients, the local model can be trained by
	\begin{equation} \label{inter}
	\mathcal{L}_{Inter}=\sum_{i \in I} \frac{-1}{\left| P(i)\right|}  \log \frac{\exp (z_i \cdot v^{y_{i}} / \tau)}{\sum_{j=1}^L\exp (z_i \cdot v^j / \tau)},
	\end{equation}
	where $y_{i}$ is the label of sample $i$. When minimizing $\mathcal{L}_{Inter}$, the embedding of each sample will get closer to the prototype of the same class while farther from the prototypes of different classes, encouraging local models to learn common attributes (\textit{i.e.} class-specific attributes) for samples with the same classes.
	
	To this end, how to produce high-quality prototypes is the key to inter-client contrastive learning. In previous studies, one common method to generate prototypes is uploading and aggregating local information. For example, Mu \textit{et al.} \cite{mu2023fedproc} and Chen \textit{et al.} \cite{chen2022personalized} uploaded features to the server directly to generate prototypes. However, it may cause privacy leakage under well-designed attacks and will introduce extra communication costs. Different from these methods, in FedIIC, we propose a new method to generate global prototypes without uploading extra information. Considering that the essence of linear classification is similarity calculation based on vector inner product, the weights of a well-trained linear classifier are nearly co-linear with the feature vectors of different classes \cite{graf2021dissecting,papyan2020prevalence,zhu2022balanced}. Therefore, the weights of a linear classifier denoted as $W=\{w^1, w^2, ..., w^L\}$, can represent the corresponding features of $L$ classes learned by the feature extractor $g(\cdot)$ to some extent. Specifically, given a global model $[f_g(\cdot), g_g(\cdot), h_g(\cdot)]$ after model aggregation in the server, the weights of $g_g(\cdot)$ are fed to $h_g(\cdot)$ to calculate the initial prototypes $\widetilde V=\{\widetilde v^1, \widetilde v^2, ..., \widetilde v^L\}$ as shown in Fig. \ref{framework}. Considering that features of different classes should have low inter-class similarity, we further fine-tune $\widetilde V$ via gradient descent by
%	To decrease inter-class similarity of $\widetilde V$, we further fine-tune $\widetilde V$ via gradient descent by
	\begin{equation} \label{finetune}
	\widetilde V \leftarrow \widetilde V - \nabla \sum_{i \in Y} \max_{j \in Y, j \neq i} (\frac{\widetilde v^i}{\left\| \widetilde v^i\right\|_2} \cdot \frac{\widetilde v^j}{\left\| \widetilde v^j\right\|_2}).
	\end{equation}
	In this way, the cosine similarity of any $({\widetilde v^i}, {\widetilde v^j})$ pair in $\widetilde V$ is minimized to be equal, resulting in $\widetilde V$ with lower inter-class similarity. This operation is called orthogonalization. Finally, the class-wise prototypes $V$ are defined as the element-wise $l_2$-normalization of $\widetilde V$ and are sent to clients for inter-client contrastive learning.
	
	\subsection{Difficulty-Aware Logit Adjustment}
	
	After calibrating the feature extractor $g(\cdot)$, one common method to calibrate the linear classifier $f(\cdot)$ is logit adjustment (LA) \cite{cao2019learning,menon2021long} to alleviate the impact of class imbalance in local training. Specifically, Zhang \textit{et al.} \cite{zhang2022federated} proposed to add per-class margins to logits and re-compute the cross entropy (CE) loss by
	\begin{equation} \label{LAloss}
	\mathcal{L}_{LA} = \sum_{i \in I} - \log \frac{{\rm exp}(f(g(x_i))_{y_i}-\delta_{y_i})}{\sum_{y^{\prime} \in \mathcal{Y}}{\rm exp}(f(g(x_i))_{y^{\prime}}-\delta_{y^{\prime}})},
	\end{equation}
	where $\delta_y$ denotes the positive per-class margin and is inversely proportional to the local class frequency $p(y)$. In this way, during local training, the logits of minority classes will increase to compensate for the item, which in turn trains the model to emphasize more on minority classes. However, the frequency-dependent margin may not be appropriate for medical data. For instance, some disease types/classes may have large intra-class variations and are difficult to diagnose even with a large amount of data, which may result in even smaller per-class margins. To address this, in FedIIC, the per-class margin is calculated based on not only the class frequency but also difficulties inspired by \cite{zhao2022adaptive}. Specifically, we define $\delta_y :=\log([\overline l_{ce}(y)]^q / p(y))$, where $\overline l_{ce}(y)$ is the average CE loss of all samples belonging to class $y$ in any round and $q$ is a hyper-parameter set as 0.25 by default. $\overline l_{ce}(y)$ is calculated as follows. At any round $r$, the total sample number of class $y$, denoted as $N^y_r$, belonging to clients of communication is first calculated. After receiving the global model from the server and before local training, each client $i$ uploads ${{l}^i_{ce}(y)}$, \textit{i.e.} the total loss of class $y$, to the server. Finally, $\overline l_{ce}(y)$ is calculated as $\frac{1}{N^y_r} \sum_i {l}^i_{ce}(y)$. This process to calculate average loss value can be privacy-preserving under the existing secure multi-party computation framework based on homomorphic encryption \cite{shen2022agnostic}. Based on the newly defined $\delta_y$, Eq. \ref{LAloss} is renamed as $\mathcal{L}_{DALA}$. Note that the calculation of $\mathcal{L}_{DALA}$ does not rely on the multi-view batch like $\mathcal{L}_{Intra}$ and $\mathcal{L}_{Inter}$. For a fair comparison with other methods trained by the CE loss, only one view of the multi-view batch is used to calculate $\mathcal{L}_{DALA}$.
	The overall loss function in local training is written as
	\begin{equation} \label{loss}
	\mathcal{L} =  \mathcal{L}_{DALA} + k_1\mathcal{L}_{Intra} + k_2\mathcal{L}_{Inter},
	\end{equation}
	where $k_1$ and $k_2$ are trade-off hyper-parameters. After minimizing $\mathcal{L}$ during the local training phase of each client, the global model is updated by FedAvg \cite{mcmahan2017communication}.

	\section{Experiments}
	
	\subsubsection{Datasets}
	\begin{figure}[!t]
		\centering
		\includegraphics[width=1\textwidth]{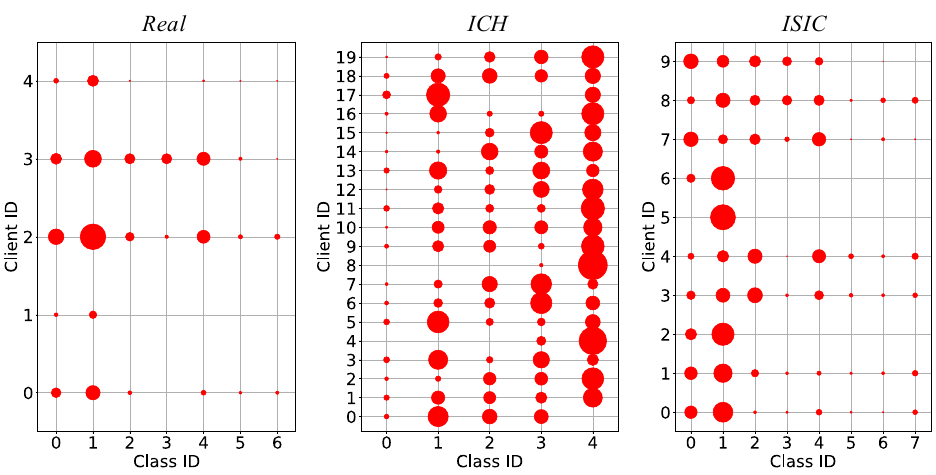}
		\caption{Illustration of imbalanced data distributions. The radius of each solid circle represents each client's data amount of a specific class.}
		\label{data}
	\end{figure}

	Three FL scenarios with class-imbalanced global data are used for evaluation, which are described as follows:
	\begin{enumerate}
		\item Real Multi-Source Dermoscopic Image Datasets (denoted as \textit{\textbf{Real}}) consisting of five data sources from three datasets, including PH$^2$ \cite{ph2}, Atlas \cite{atlas}, and HAM10000 \cite{tschandl2018ham10000} where each source is treated as an individual client. For evaluation, we construct a separate test set by randomly sampling from the training set of ISIC 2019 \cite{tschandl2018ham10000,combalia2019bcn20000} and ensure that the test set has no overlap with the above five data sources.
		
		\item Intracranial Hemorrhage Classification (denoted as \textit{\textbf{ICH}}). The RNSA ICH dataset \cite{flanders2020construction}, containing five ICH subtypes, is adopted for experiments. The same pre-processing strategies in \cite{jiang2022dynamic,liu2021federated} are adopted, and images with only one single hemorrhage type are selected. Following \cite{jiang2022dynamic,liu2021federated}, data is split according to 7:1:2 for training, validation, and testing respectively. To simulate heterogeneous multi-source data, following \cite{shang2022federated}, Dirichlet distribution, \textit{i.e.} $Dir(\alpha=1.0)$, is used to divide the training set to 20 clients.
		
		\item Skin Lesion Classification (denoted as \textit{\textbf{ISIC}}).The training data of ISIC 2019 \cite{tschandl2018ham10000,combalia2019bcn20000}, containing eight classes, is used for evaluation. Following \cite{jiang2022dynamic,liu2021federated}, we split the dataset by 7:1:2 for training, validation, and testing respectively. Similarly, Dirichlet distribution, \textit{i.e.} $Dir(\alpha=1.0)$, is used to generate highly heterogeneous data partitions of 10 clients.
	\end{enumerate}
	Data distributions of the three training settings are illustrated in Fig. \ref{data}, and imbalance ratios are 35.43, 19.59 and 57.60, respectively.
	
	\subsubsection{Implementation Details} \label{details}
	\begin{table}[!t]
		\centering
		\caption{Quantitative comparison results under the \textit{Real}, \textit{ISIC}, and \textit{ICH} settings. For \textit{Real}, the average results (\%) from the last five rounds are reported. For \textit{ISIC} and \textit{ICH}, the results (\%) based on the best model (evaluated by the validation set) on the testing set are reported. The best results are marked in bold.}\label{compare}
		\resizebox{\linewidth}{!}{
			\begin{tabular}{c|c|ccc|ccc|ccc}
				\toprule
				\hline
				\multirow{3}{*}{Methods} & \multirow{3}{*}{Year} & \multicolumn{9}{c}{Datasets}                                                                                                                                                                     \\ \cline{3-11} 
				&                         & \multicolumn{3}{c|}{\textit{Real}}                                    & \multicolumn{3}{c|}{\textit{ISIC}}                                    & \multicolumn{3}{c}{\textit{ICH}}                 \\ \cline{3-11} 
				&                         & BACC           & F1      & \multicolumn{1}{c|}{ACC}            & BACC           & F1             & \multicolumn{1}{c|}{ACC}            & BACC           & F1             & ACC           \\ \hline
				FedAvg \cite{mcmahan2017communication}              & AISTATS'17              & 45.21          & 44.47          & 44.57    & 49.41          & 54.31          & 72.50          & 73.75          & 77.35          & 84.83           \\
				FedProx \cite{li2020federated}             & MLSys'20                & 45.61          & 44.90          & 44.89       & 69.00          & 69.46          & 80.50       & 79.62          & 82.45          & 86.78          \\
				MOON \cite{li2021model}               & CVPR'21                 & 44.40          & 43.28          & 43.68      & 66.31          & 71.27          & 81.38          & 77.05          & 78.81          & 84.87           \\
				FedProc \cite{mu2023fedproc}             & FGCS'23                 & 38.83          & 37.98          & 39.36    & 31.16          & 35.45          & 66.88           & 73.29          & 76.23          & 84.89          \\
				FedRS \cite{li2021fedrs}               & KDD'21                  & 45.23          & 44.50          & 44.46      & 24.93          & 26.01          & 61.39        & 72.44          & 76.51          & 84.13         \\
				FedLC \cite{zhang2022federated}               & ICML'22                 & 46.73          & 45.88          & 45.60       & 45.84          & 41.89          & 70.33       & 76.53          & 78.96          & 84.92           \\
				FedFocal \cite{sarkar2020fed}            & IJCAI'20                & 44.00          & 43.31          & 42.96            & 47.68          & 38.29          & 56.99       & 63.04          & 54.80          & 52.30          \\
				PRR-Imb \cite{chen2022personalized}             & TMI'22                  & 50.49          & 47.60          & 47.48     & 49.97          & 46.52          & 68.18       & 71.72          & 69.98          & 78.85          \\
				CLIMB \cite{shen2022agnostic}               & ICLR'22                 & 46.07          & 45.91          & 45.86         & 49.70          & 52.32          & 71.65       & 72.64          & 76.08          & 84.73          \\
				CReFF \cite{shang2022federated}                & IJCAI'22                & 51.13          & 48.56          & 49.46      & 71.52          & 57.83          & 72.92   & 82.21          & 74.64          & 81.63         \\ \hline
				FedIIC (ours)              & -                       & \textbf{55.12} & \textbf{51.57} & \textbf{51.67}  & \textbf{78.84} & \textbf{78.05} & \textbf{85.71} & \textbf{84.22} & \textbf{84.73} & \textbf{87.77}  \\ \hline
				\bottomrule
			\end{tabular}
		}
	\end{table}
	EfficientNet-B0  \cite{tan2019efficientnet}, pre-trained by ImageNet \cite{deng2009imagenet}, is adopted as the backbone trained by an Adam optimizer with betas as 0.9 and 0.999, a weight decay as 5e-4, constant learning rates of 1e-4 for \textit{Real} and 3e-4 for both \textit{ICH} and \textit{ISIC}, and a batch size of 32. For \textit{ICH}, the multi-view batch for contrastive learning is generated by following \cite{jiang2022dynamic,liu2021federated}. For both \textit{Real} and \textit{ISIC}, the multi-view batch is generated by 1) RandAug \cite{cubuk2020randaug} and 2) SimAugment \cite{chen2020simple}. The hyper-parameters $k_1$ and $k_2$ in Eq. \ref{loss} are set as 2.0. For federated training, the local training epoch is set as 1 and the global training round is set as 200 for \textit{ICH} and \textit{ISIC} and 30 for \textit{Real}. At each round, all clients (\textit{i.e.}, 100\%) are included for model aggregation. 
	
	\subsection{Comparison with State-of-the-art Methods}
	
	Ten related approaches are included for comprehensive comparison, including FedAvg \cite{mcmahan2017communication}, FedProx \cite{li2020federated} addressing data heterogeneity, MOON \cite{li2021model} and FedProc \cite{mu2023fedproc} utilizing contrastive learning in FL, FedFocal \cite{sarkar2020fed} utilizing focal loss \cite{lin2017focal} for balancing, FedRS \cite{li2021fedrs} addressing the class-missing problem, FedLC \cite{zhang2022federated} applying frequency-dependent logits adjustment in FL, PRR-Imb \cite{chen2022personalized} training personalized models with heterogeneous and imbalanced data, and CLIMB \cite{shen2022agnostic} and CReFF \cite{shang2022federated} addressing class-imbalance global data in FL. All the methods share the same experimental details described above for a fair comparison. \textit{More implementation details and visualization results can be found in supplemental materials.}
	% Note that for data augmentation, the stronger augmentation to generate the multi-view batch is employed for all comparison methods, hence our method does not introduce stronger augmentation for performance gain, which is fair.
	
	Following the ISIC 2019 competition, balanced accuracy (BACC) is used as the primary metric for class-imbalanced testing sets. Two key metrics in classification, \textit{i.e.} F1 score (F1) and accuracy (ACC) are also employed for evaluation. Comparison results are summarized in Tab. \ref{compare}. As can see, FedIIC achieves the best performance against all previous methods across the three metrics, outperforming the second-best approach (CReFF) by 3.99\%, 7.32\%, and 2.01\% in BACC on \textit{Real}, \textit{ISIC}, and \textit{ICH} respectively.
	
	\subsection{Ablation Study}
	
	To validate the effectiveness of each component in FedIIC, a series of ablation studies are conducted on \textit{ISIC} and \textit{ICH} following the same experimental details described in Section \ref{details}. Quantitative results are summarized in Tab. \ref{abalation}. Under severe global imbalance, FedAvg is struggling. With the introduction of DALA, the performance is improved in BACC but degraded in F1. It is consistent with the quantitative results between CReFF and FedAvg on \textit{ICH} in Tab. \ref{compare}, indicating the limitation of only eliminating class bias through classifier calibration while ignoring attribute bias. The above results validate the necessity of addressing the imbalance in feature learning for performance improvement. Therefore, introducing either intra- or inter-client contrastive learning for better representation learning under class imbalance is beneficial in both BACC and F1. By combining all the components, FedIIC achieves the best overall performance, outperforming FedAvg with large margins.
	
	Ablation studies of hyper-parameters in FedIIC are conducted on \textit{ISIC} as stated in Tab. \ref{param}. Setting $t=0$ encounters noticeable performance degradation, indicating the necessity of dynamic temperatures based on class priors in intra-client contrastive learning. Meanwhile, the performance gap between the initial prototypes $\widetilde V$ with and without orthogonalization validates the effectiveness of reducing inter-class similarity in prototypes. When introducing difficulty to logit adjustment (\textit{i.e.}, $d=0.25$), we observe an increase in BACC and a decrease in F1, which is consistent with the above analysis in Tab. \ref{compare} (\textit{i.e.}, CReFF vs. FedAvg).
	\begin{table}[!t]
		\centering
		\begin{minipage}{0.6\linewidth}
			\centering
			\caption{Component-wise study.}\label{abalation}
			\begin{tabular}{cccc|cc|cc}
				\toprule
				\hline
				\multirow{2}{*}{FedAvg} & \multirow{2}{*}{DALA} & \multirow{2}{*}{Intra} & \multirow{2}{*}{Inter} & \multicolumn{2}{c|}{\textit{ISIC}}       & \multicolumn{2}{c}{\textit{ICH}}         \\ \cline{5-8} 
				&                       &                        &                        & BACC           & F1             & BACC           & F1             \\ \hline
				\checkmark                       &                       &                        &                        & 49.41          & 54.31          & 73.75          & 77.35          \\
				\checkmark                       & \checkmark                     &                        &                        & 50.65          & 42.12          & 81.26          & 76.47          \\
				\checkmark                       & \checkmark                     & \checkmark                      &                        & 51.30          & 43.84          & 81.96          & 82.68          \\
				\checkmark                       & \checkmark                     &                        & \checkmark                      & 75.78          & 76.55          & 83.81          & 82.39          \\
				\checkmark                       & \checkmark                     & \checkmark                      & \checkmark                      & \textbf{78.84} & \textbf{78.05} & \textbf{84.22} & \textbf{84.73} \\ \hline
				\bottomrule
			\end{tabular}
		\end{minipage}
		\begin{minipage}{0.39\linewidth}
			\centering
			\caption{Parameter-wise study.}\label{param}
			\begin{tabular}{c|c|cc}
				\toprule
				\hline
				\multicolumn{2}{c|}{Param}                        & BACC           & F1             \\ \hline
				\multirow{2}{*}{$t$}             & 0.0            & 76.33          & 75.43          \\ 
				& 0.5            & \textbf{78.84} & \textbf{78.05}          \\ \cline{1-4} 
				\multirow{2}{*}{$orth.$}         & w/o            & 72.64          & 75.34          \\
				& w              & \textbf{78.84} & \textbf{78.05}          \\ \cline{1-4} 
				\multirow{2}{*}{$d$}             & 0.0            & 77.32          & \textbf{78.41}          \\
				& 0.25           & \textbf{78.84} & 78.05 \\ \hline
				\bottomrule
			\end{tabular}
		\end{minipage}
	\end{table}
	
	\section{Conclusion}
	This paper discusses a more realistic federated learning (FL) setting in medical scenarios where global data is class-imbalanced and presents a novel framework FedIIC. The key idea behind FedIIC is to calibrate both the feature extractor and the classification head to simultaneously eliminate attribute biases and class biases. Specifically, both intra- and inter-client contrastive learning are introduced for balanced feature learning, and difficulty-aware logit adjustment is deployed to balance decision boundaries across classes. Experimental results on both real-world and simulated medical FL scenarios demonstrate FedIIC's superiority against the state-of-the-art FL approaches. We believe that this study is helpful to build real-world FL systems for clinical applications.
	\\
	\\
	\textbf{Acknowledgement.} This work was supported in part by the National Natural Science Foundation of China under Grants 62202179 and 62271220, in part by the Natural Science Foundation of Hubei Province of China under Grant 2022CFB585, and in part by the Research Grants Council GRF Grant 16203319. The computation is supported by the HPC Platform of HUST.

%	\newpage
	\bibliographystyle{splncs04}
	\bibliography{main}
\end{document}